\documentclass[preprint,12pt]{elsarticle}

\usepackage{amsmath,amssymb,amsfonts}
\usepackage{algorithmic}
\usepackage{graphicx}
\usepackage{textcomp}
\usepackage{xcolor}
\def\BibTeX{{\rm B\kern-.05em{\sc i\kern-.025em b}\kern-.08em
    T\kern-.1667em\lower.7ex\hbox{E}\kern-.125emX}}

\usepackage{multirow}
\usepackage{hhline}
\usepackage{comment}
\usepackage[colorlinks=false, linkcolor=blue, citecolor=blue, bookmarks=true,pagebackref=false]{hyperref}
\usepackage{hyphenat}
\usepackage{subcaption}
\usepackage{float}
\usepackage{fancybox}
\usepackage{array,booktabs}
\usepackage[T1]{fontenc}
\usepackage{listings}
\usepackage[normalem]{ulem}
\usepackage{natbib}

\newcommand{\rqbox}[1]{
\begin{center}
\vspace{-0.1cm}
\cornersize{.2}
\setlength{\fboxsep}{7pt}
\ovalbox{\begin{minipage}{5.2in}
{ #1}
\end{minipage}}
\vspace{-0.1cm}

\end{center}}

\newcommand{\sw}[1]{\textcolor{green}{{\it [Shaowei says: #1]}}}

\newcommand{\rqone}{RQ1: How prevalent is the information highlighting in SO answers?}
\newcommand{\rqtwo}{RQ2: What types of information are highlighted with \emph{Code} formatting in SO answers?}
\newcommand{\rqthree}{RQ3: What types of information are highlighted with \emph{Bold}, \emph{Italic}, \emph{Delete}, and \emph{Heading} formatting in SO answers?}
\newcommand{\rqfour}{RQ4: Can we recommend the highlighted content in SO answers automatically?} 

\makeatletter
\def\ps@pprintTitle{%
 \let\@oddhead\@empty
 \let\@evenhead\@empty
 \def\@oddfoot{}%
 \let\@evenfoot\@oddfoot}
\makeatother
\begin{document}

\begin{frontmatter}

\title{Studying and Recommending Information Highlighting in Stack Overflow Answers}


\author[1]{Shahla Shaan Ahmed}

\ead{ahmeds27@myumanitoba.ca}

\affiliation[1]{organization={Department of Computer Science, University of Manitoba},
country={Canada}}

\author[1]{Shaowei Wang}

\ead{shaowei.wang@umanitoba.ca}

\author[2]{Yuan Tian}

\ead{y.tian@queensu.ca}


\affiliation[2]{organization={School of Computing, Queen's University},
country={Canada}}

\author[3]{Tse-Hsun (Peter) Chen}

\ead{peterc@encs.concordia.ca}


\affiliation[3]{organization={Department of Computer Science and Software Engineering, Concordia University},
country={Canada}}

\author[4]{Haoxiang Zhang}

\ead{haoxiang.zhang@acm.org}


\affiliation[4]{organization={Huawei, Canada},
country={Canada}}

\begin{abstract}
\textbf{Context:} Navigating the knowledge of Stack Overflow (SO) remains challenging. To make the posts vivid to users, SO allows users to write and edit posts with Markdown or HTML so that users can leverage various formatting styles (e.g., bold, italic, and code) to highlight the important information. Nonetheless, there have been limited studies on the highlighted information. \\  
\textbf{Objective:} We carried out the first large-scale exploratory study on the information highlighted in SO answers in our recent study. To extend our previous study, we develop approaches to automatically recommend highlighted content with formatting styles using neural network architectures initially designed for the Named Entity Recognition task. \\
\textbf{Method:} In this paper, we studied 31,169,429 answers of Stack Overflow. For training recommendation models, we choose CNN-based and BERT-based models for each type of formatting (i.e., Bold, Italic, Code, and Heading) using the information highlighting dataset we collected from SO answers. \\
\textbf{Results:} Our models achieve a precision ranging from 0.50 to 0.72 for different formatting types. It is easier to build a model to recommend Code than other
types. Models for text formatting types (i.e., Heading, Bold, and Italic) suffer low recall. Our analysis of failure cases indicates that the majority of the failure cases are due to missing identification. One explanation is
that the models are easy to learn the frequent highlighted words while struggling to learn less frequent words (i.g., long-tail knowledge).\\
\textbf{Conclusion:} Our findings suggest that it is possible to develop recommendation models for highlighting information for answers with different formatting styles on Stack Overflow.\\
\end{abstract}


\begin{keyword}

Stack Overflow \sep Information highlighting \sep Named entity recognition \sep Deep learning

\end{keyword}

\end{frontmatter}



\section{Introduction}\label{sec:introduction}
Technical question and answer (Q\&A) sites such as Stack Overflow (SO) have become increasingly important for software developers to share knowledge and contribute to communities. Despite Stack Overflow's success and prevalence, navigating the knowledge on it remains challenging~\cite{gottipati2011finding,nadi2020essential,xu2017answerbot,zhang2021comments}. The previous study shows that finding answers in long posts remains one of the challenges~\cite{nadi2020essential,xu2017answerbot}. 37\% of all questions on Stack Overflow have more than one answer, and the average length of an answer is 789 characters~\cite{nadi2020essential}. To make the posts vivid to users, Stack Overflow platform allows users to edit their posts with Markdown and HTML~\cite{MarkdownHelp,StackExchangeMarkdown}, so that users can leverage various formatting styles (e.g., \textbf{Bold}, \emph{Italic}, and \colorbox{lightgray}{Code}) to highlight text and direct other users' attention toward the most important information within posts.

Information highlighting has demonstrated its effectiveness in various domains~\cite{nguyen2015combining,ramirez2019understanding,strobelt2015guidelines,wilson2016crowdsourcing,wu2003improving}, including in Software Engineering~\cite{sarkar2015impact} (e.g., it saves the reading time of humans and helps programming novices with code comprehension with syntax highlighting). However, little is known about how information is highlighted on technical Q\&A sites (e.g., Stack Overflow). For example, how prevalent is the information highlighted? What content is highlighted using different formatting styles? Understanding this could provide a landscape of the usage of information highlighting on technical Q\&A sites and shed light on the information that is considered important to developers.

\textbf{Previous study:} In our previous study~\cite{ahmed2022first}, we performed the first large-scale study on the five most commonly used information highlighting types, which are \emph{Bold}, \emph{Italic}, \emph{Code}, \emph{Delete}, and \emph{Heading}, to understand their characteristics and what information is highlighted in the text description of SO answers with them\footnote{Note that an SO answer may contain both text and code block, we focus on the information highlighting in text description.}. Our analysis of 55,209,643 information highlighting instances across 14,845,929 answers on Stack Overflow reveals that: Overall, information highlighting is prevalent on SO, i.e., 47.6\% (14,845,929 out of 31,169,429) of the answers use the studied formatting types to highlight information. 38.5\%, 11.3\%, and 7.2\% of the answers use \emph{Code}, \emph{Bold}, and \emph{Italic}, respectively, and their highlighted content is short (median length is one word). \emph{Code} formatting is mainly used to highlight source code content, such as identifiers (63.5\%). \emph{Code} is also used to highlight content other than source code, such as Software (4.9\%) and Equation (5.2\%). \emph{Italic} and \emph{Bold} are frequently used to highlight source code content, as well as content that warns about the context where the provided solution works or does not work, updates on answers, and references to an internal or external resource. 

\textbf{Extention:} Since information highlighting is prevalent on SO, however, identifying the appropriate content to highlight can be a challenging task for SO users (typically for new users), as they must properly pinpoint the focal points and select an appropriate formatting style, which can be time-consuming and require expertise. Recommending important information to highlight in a post can help improve the presentation of the post and make the important information more visible to SO users. Therefore, in this paper, we extend our recent study~\cite{ahmed2022first} by investigating the potential of recommending highlighted content automatically. To do so, we adapt two models (i.e., CNN and BERT) to automatically identify highlighted content for each type of formatting. More specially, we train NER models on each formatting (i.e., Heading, Bold, Italic, and Code) type and recommend the content to highlight. CNN-based models outperform BERT-based models. CNN models achieve precision values ranging from 0.50 to 0.72. The CNN models for automatic source code content highlighting achieve a recall score of 0.66 and an F1 score of 0.69, outperforming the models for other formatting types. In addition, CNN-based models outperform BERT-based models.
By analyzing the failure cases, we observed that the majority of the failure cases are due to missing identification (i.e., the model misses the content that is supposed to be highlighted). Our analysis shows that the models tend to learn the frequent highlighted words while struggling to learn less frequent words (i.g., long-tail knowledge). We make our replication package public at \url{https://github.com/shaoweiwang2010/REP_2022_Information_highlight_SO}.


\vspace{0.1cm}
\noindent \textbf{Paper Organization.} Section~\ref{sec:backrelated} presents the background and motivation of the work. Section~\ref{sec:data} explains our research questions and how we prepare our data to answer research questions. Section~\ref{sec:method} introduces the methodology of our study. In Section~\ref{sec:results}, we present the results of our research questions. Section~\ref{sec:disc} discusses the implications of our study and threats to validity. Section~\ref{sec:related} presents the related work. Finally, Section~\ref{sec:conclusion} concludes our study.

\section{Background}\label{sec:backrelated}
\subsection{Background}\label{sec:background}

\begin{figure}[h]
    \centering
    \includegraphics[width=0.9\columnwidth]{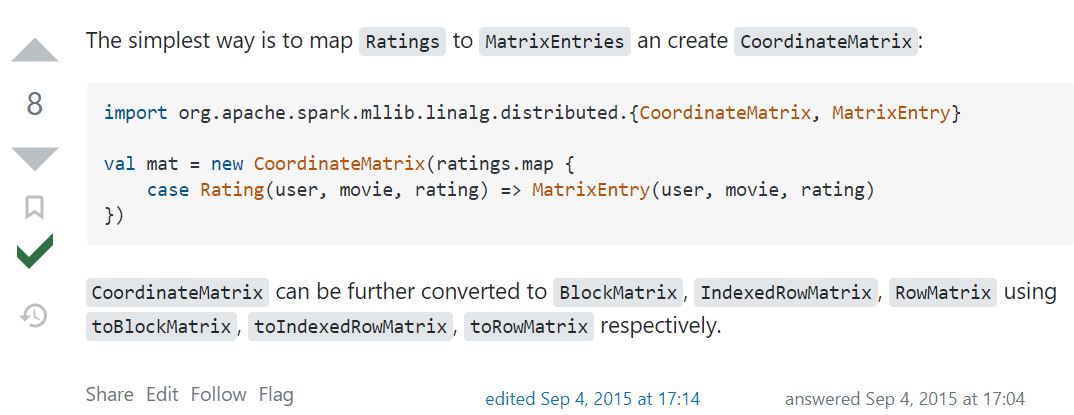}
    \caption{An example of SO answers in which the code-related content is highlighted.}
    \label{fig:example}
\end{figure}

Stack Overflow allows users to use Markdown and HTML to write and edit posts~\cite{MarkdownHelp,StackExchangeMarkdown}. Certain formatting types are used to highlight the information in text description with Markdown and HTML tags. In this study, we focus on the five most commonly used formatting types, which are \emph{Bold}, \emph{Italic}, \emph{Delete}, \emph{Code}, and \emph{Heading}. 
In Table~\ref{tab:formatTypes}, we present the HTML tags and their equivalent Markdown syntax for each formatting type. We group HTML tags and Markdown formatting based on their rendering effect. For example, we group <b> and <strong> as \emph{Bold} since they render the same effect in the browser when users read the posts on Stack Overflow. In this study, we consider a highlighted sequence of words as an \textit{highlighted instance}. One post could have multiple highlighted instances. For instance, in Figure~\ref{fig:example}, the post has 10 Code instances\footnote{\url{https://stackoverflow.com/questions/32402475}}.

In the rest of this paper, we refer to the formatting types \emph{Bold}, \emph{Italic}, \emph{Heading}, and \emph{Delete} as \emph{Text} formatting for simplicity's sake. We use \emph{Code} and \emph{Code} formatting, \emph{Text} and \emph{Text} formatting exchangeably.


\begin{table}
\caption{The studied formatting types that are used to highlight information, their corresponding HTML tags, and equivalent Markdown. HTML tags are grouped based on their rendering effects.}\label{tab:formatTypes}
\begin{tabular}{|p{2.5cm}|p{3cm}|p{6.7cm}|}
  \hline
 \textbf{Type} & \textbf{HTML tags} & \textbf{Equivalent Markdown }\\
 \hline
 Code & <code> & 'example' \\
   \hline
 Bold  & <b>, <strong> & **example**, \_\_example\_\_  \\
 \hline
 Italic & <i>, <em> &  *example*  \\
 \hline
 Delete & <del>, <s> & None \\
  \hline
  Heading & <h1>, <h2>, <h3>, <h4>, <h5>, <h6> & \# example, \#\# example, \#\#\# example, \#\#\#\# example, \#\#\#\#\# example, \#\#\#\#\#\# example \\
  \hline
\end{tabular}
\vspace{-0.2in}
\end{table}

\subsection{Motivation}\label{sec:motivation}
\begin{figure}
    \centering
    \includegraphics[width=\linewidth]{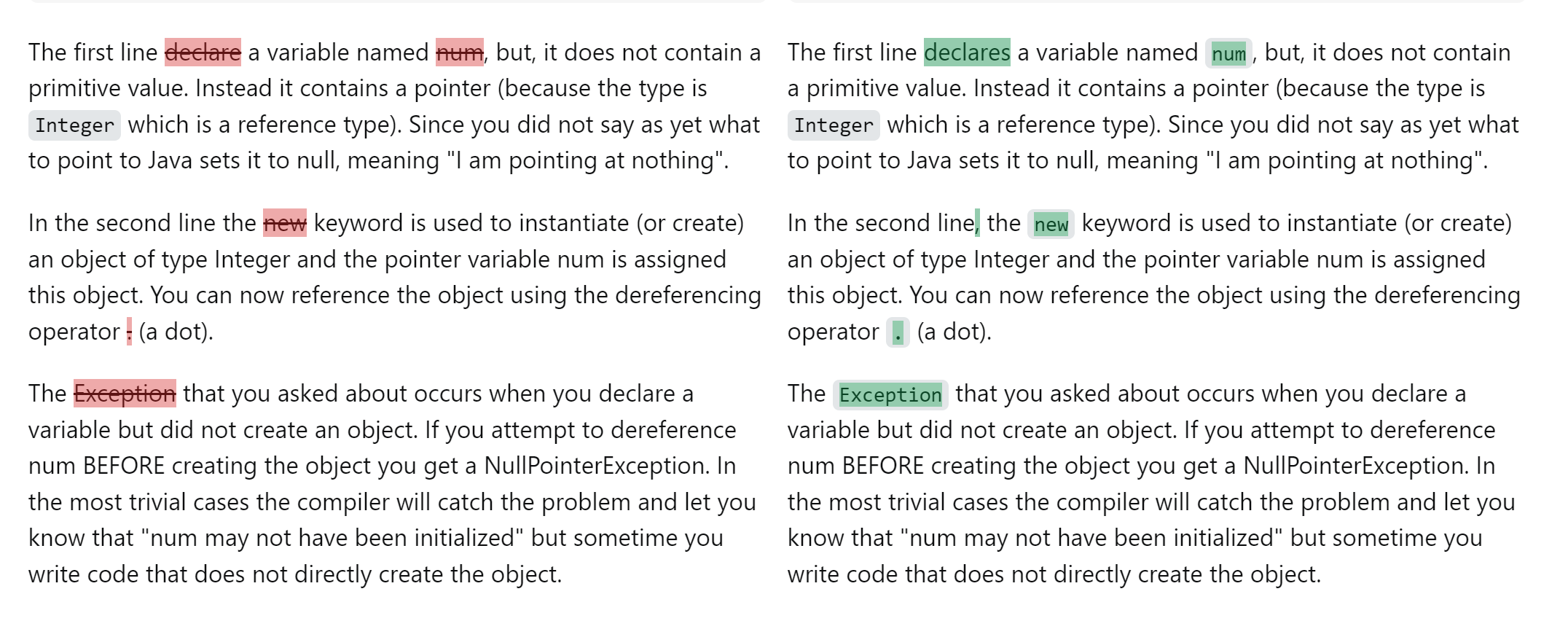}
    \caption{An example of revising highlight to improve the presentation of the post.}
    \label{fig:revisionExample}
\end{figure}

As discussed in Section~\ref{sec:backrelated}, Stack Overflow allows users to apply different types of formatting on content to highlight the information. By comprehending the usage practices of those formatting types, we can provide insights into what information is important from users' perspectives and provide insights for future research. For instance, we can provide insight into the downstream research that leverages SO information to facilitate software engineering tasks (e.g., API documentation enrichment\cite{li2018improving,treude2016augmenting}). Furthermore, identifying the appropriate content to highlight can be a challenging task for SO users, as they must properly pinpoint the focal points and select an appropriate formatting style, which can be time-consuming and require expertise. For instance, we observe that, in many SO answer posts, certain contents were not initially highlighted in the first version of the post but were highlighted in subsequent versions, such as the second or third. For instance, as an example shown in Figure~\ref{fig:revisionExample}, in the second and third revisions of a post, a user highlighted the code-related content (e.g., ``Integer'', ``Exception'') to improve the presentation of the post. In addition, we observed (see details in Sections~\ref{sec:rq2} and \ref{sec:rq3}). Therefore, in this study, we also aim to investigate the feasibility of building automatic approaches to recommend content to be highlighted with formatting types. 

\section{Experimental settings}\label{sec:data}

\subsection{Research questions}

This study aims to understand how information highlighting is used in SO answers and what content is highlighted using different formatting types. Hence, we formulate our study by answering the following three research questions:

\begin{itemize}
  \item \rqone{}
  \item \rqtwo{}
  \item \rqthree{}
  \item \rqfour{} 
\end{itemize}


In RQ1, we aim to understand the prevalence of the usage of the studied highlighting formatting and their characteristics. We study \emph{Code} formatting and the rest formatting types (i.e., \emph{Bold}, \emph{Italic}, \emph{Heading}, and \emph{Delete}) in separated RQs (RQ2 and RQ3) due to their different nature. We study the information highlighted by different formatting types. In RQ4, we explore the feasibility of utilizing machine learning models to recommend which information to highlight.

\subsection{Data preparation}
To answer the four RQs, we downloaded a data dump of Stack Overflow from the Stack Exchange data dump dated March 2021\footnote{https://archive.org/details/stackexchange}. The data dump contains details information about posts (i.e., questions and answers), as well as their revision history. In this study, we include all the answers and we ended up with 31,169,429 answer posts.


As the dataset was large for processing, we imported it into MySQL. For each answer post, we first extract the textual content from its body and exclude code block(s). Note that we extracted code block(s) using tag ``<pre>''. Next, we apply the regular expressions defined in Table~\ref{tab:formatTypes} to identify information highlighting in each answer post, using the Python library \emph{Regex}. In the end, we extract the content that is highlighted by the studied formatting types. 

Table~\ref{tab:Datasetmodel} presents the basic statistics of our dataset for different formatting types.

\section{Methodology}\label{sec:method}

\begin{figure}
    \centering
    \includegraphics[width=1\linewidth]{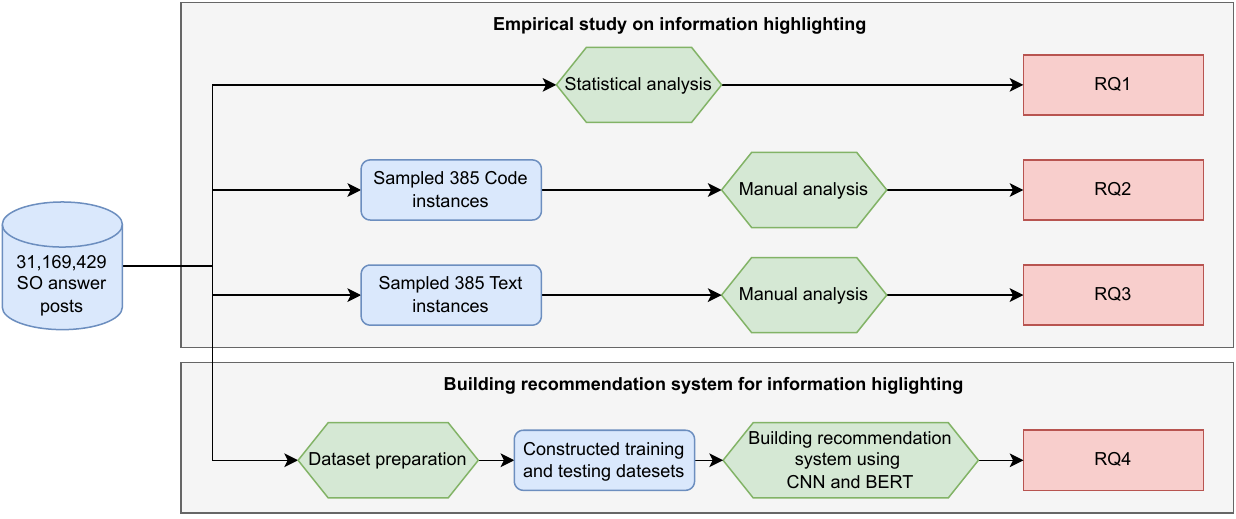}
    \caption{Framework of our study.}
    \label{fig:framework}
\end{figure}

Figure~\ref{fig:framework} presents an overview of the methodology for our study. Particularly, our study could be decomposed into two parts. In part 1, we perform an empirical study on the information highlighting of SO posts. In part 2, we investigate the feasibility of building a recommendation system for information highlighting. We elaborate on the details of the approach to answering RQs below. 

\subsection{Approach for RQ1}
To understand the prevalence of information highlighting in SO answers, we compute the percentage of answers that have the studied formatting types (see Table~\ref{tab:formatTypes}) and the percentage of words highlighted in each answer with each formatting. In addition, we compute the basic statistics of information highlighting instances to understand the characteristics of each formatting type. More specifically, we calculate the distribution of instances for each formatting and the number of words highlighted with them. 

\subsection{Approach for RQ2}
In this RQ, we aim to understand what content is highlighted with the \emph{Code}. First, we randomly sampled 385 \emph{Code} instances with a 5\% interval and 95\% confidence level. Since there is no existing terminologies we could reuse for this purpose, we manually performed a lightweight open coding-like process~\cite{seaman1999qualitative,zhang2019empirical} to identify the type of content highlighted with \emph{Code}. The process involves three phases and is performed by the first authors (A1 and A2) of the paper. In phase I, A1 and A2 derived a draft list of types based on 50 \emph{Code} instances. During the phase, the types were revised and refined. In phase II, A1 and A2 independently applied the derived types to the rest 335 samples. They took notes regarding the diffidence or ambiguity during the labeling. During this phase, no new types were introduced. In phase III, A1 and A2 discussed the results from Phase II to resolve any disagreements until a consensus was reached. The coding process has a Cohen’s kappa of 0.8 (measured before starting Phase III), which indicates a substantial level of the inter-rater agreement~\cite{viera2005understanding}. Table~\ref{tab:codelabel} presents the definition and the corresponding example of the derived types for \emph{Code}.

\subsection{Approach for RQ3}
In this RQ, we aim to understand what content is highlighted with the studied \emph{Text} formatting (i.e., \emph{Bold}, \emph{Italic}, \emph{Delete}, and \emph{Heading}). Similar to RQ2, we performed an manual study. We randomly sampled 385 \emph{Text} instances with a 5\% interval and 95\% confidence level. We ended up with 177 \emph{Bold}, 169 \emph{Italic}, 3 \emph{Delete}, and 36 \emph{Heading} instances.
We then identified the types of highlighted content by following the methodology used in RQ2. The coding process has a Cohen’s kappa of 0.78 (measured before starting Phase III), which indicates a substantial level of the inter-rater agreement~\cite{viera2005understanding}. Table~\ref{tab:textlabel} presents the definition and the corresponding example of the derived types for \emph{Text} formatting.

\subsection{Approach for RQ4}\label{sec:rq4approach}
In the previous RQs, we focused on analyzing the highlighted content on Stack Overflow and we noticed that users misuse formatting types. For instance, users used \textit{Bold} and \textit{Italic} to highlight code. Therefore, we aim to explore the feasibility of implementing an automated machine learning-based method for content highlighting. This feature would be helpful, particularly for SO users who are inexperienced in the practice of highlighting information and seek to automatically highlight critical content when composing their posts.
Specifically, we aim to create a model that can highlight information in an input answer post using appropriate formatting types (Bold, Italic, Delete, Code, and Heading). To achieve this, we use neural network architectures initially designed for Named Entity Recognition (NER) without actually training NER models. Our task shares similarities with NER, such as the short length of labeled (highlighted) tokens and a limited number of text labeling (formatting) options. Moreover, both NER and automated information highlighting can be formed as a sequence labeling task. Note that we train the neural networks from scratch on our information highlighting dataset.


\begin{figure}
    \centering
    \includegraphics[scale=0.55]{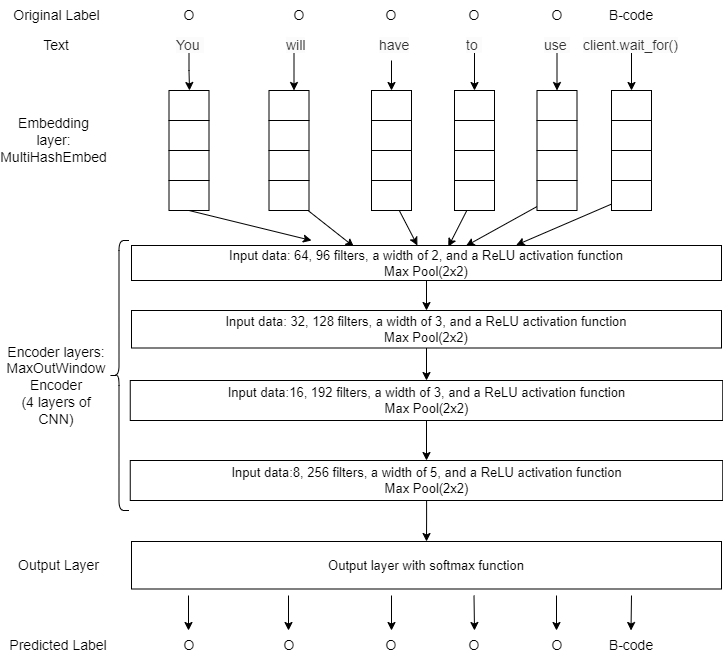}
    \caption{Layers of the CNN model}
    \label{fig:NER_layers}
\end{figure}

In this RQ, we have used two different structured models to investigate. One model is based on the Convolutional Neural Network (CNN) model and another is based on the transformer model, BERT~\cite{devlin2019bert}. We selected CNN because it has been widely used in NER tasks to capture the context information for predicting NER labels and achieved good performance~\cite{chiu2016named,zhu2018gram,jehangir2023survey}. Compared to Recurrent Neural Network-based approaches (RNN), CNN-based approaches are more efficient. In recent years, the emergence of pre-trained models, such as BERT~\cite{devlin2019bert}, T5~\cite{t5}, GPT~\cite{brown2020language} have re-evolute the landscape of the natural language processing (NLP) domain and demonstrated their effectiveness in NLP tasks, including NER tasks. Pre-trained models encapsulate knowledge gleaned from a massive amount of training data, bypassing the need for training from scratch, and could be easily adapted to downstream tasks by fine-tuning. BERT is an encoder-based model and considers the contexts of both preceding and subsequent words. BERT has been proven effective in embedding the context for NER task~\cite{souza2019portuguese,chang2021chinese,hakala2019biomedical}. Therefore, we also investigate BERT in this RQ.

Figure~\ref{fig:NER_layers} presents the architecture of our CNN model. In this paper, we use the lib spaCy~\cite{spacyNER} as the implementation of CNN model. We opt to utilize spaCy since it has emerged as the de facto standard for practical NLP due to its speed, robustness, and performance~\cite{le2022taxonerd}. The model architecture comprises a token embedding layer (128 dimensions), four convolutional layers, and an output layer. The token embedding layer is named as \textit{MultiHashEmbed}. It features an embedding layer that embeds a number of lexical attributes separately. Then the results are concatenated as a vector semantic representation of a token for passing through a feed-forward network. As it uses hashing to store embeddings in memory, it also supports faster processing of large datasets. The embedding layer is followed by a 4-layered Convolutional Neural Network (CNN), named \textit{MaxoutWindowEncoder}. Each of the convolutional layers is followed by a max pooling layer. The final layer is a softmax layer for predicting labels on tokens. Training employs cross-entropy loss and the Adam optimizer. 

\begin{figure}
    \centering
    \includegraphics[scale=0.55]{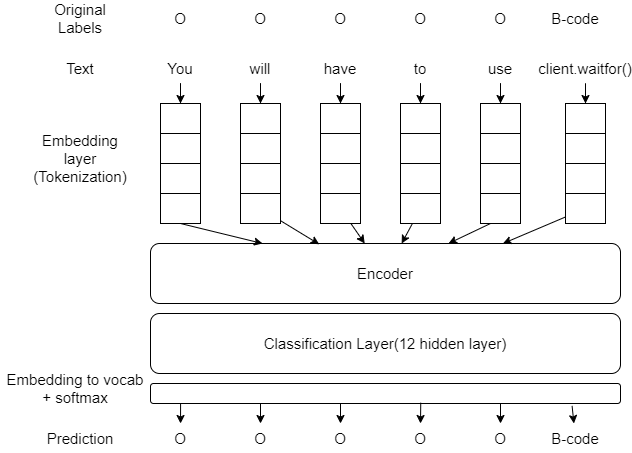}
    \caption{Layers of the NER BERT model}
    \label{fig:NER_layers_bert}
\end{figure}

Figure~\ref{fig:NER_layers_bert} presents the architecture of BERT model, which consists of multiple transformer blocks, each containing a self-attention mechanism for getting the context of the input sequence. In this work, we fine-tuned BERT for our prediction task. We used the most recent checkpoint of BERT model from HuggingFace platform~\cite{HuggingFace}.

As our HTML highlighting tags were categorized into five distinct types (see Section~\ref{sec:background} for details), we initially attempted to train a universal model that would accommodate all five types. However, this approach yielded unsatisfactory results, as the model tended to highlight words as code by default while ignoring other formatting types. Consequently, we trained separate models for each highlighting type. Our findings in RQ1 showed that the Delete type was scarcely used in comparison to the other types, leading us to omit it from our analysis in RQ4. Therefore, we train four models, i.e., $Model_{Bold}$, $Model_{Italic}$, $Model_{Heading}$, and $Model_{Code}$ with each of CNN model and BERT models.


\vspace{0.1cm}
\noindent\underline{Dataset preparation:} 
Prior to constructing the training and testing datasets, we carefully cleaned the dataset to eliminate misuse instances (e.g., using bold formatting to highlight code) to ensure that the dataset used for model training would accurately represent the intended formatting of information. To achieve this, we implemented the following pre-processing steps to clean the data.

\begin{itemize}
    \item \textbf{Cleaning up code dataset.} According to Stack Overflow guidelines, users are encouraged to highlight code content using the \textit{code} tag exclusively. However, our analysis of RQ3 results revealed instances where \textit{code} tag was improperly used for Path, software, Terminology, and Equation. Therefore, our first task is to remove such cases from our code dataset. We used regular expressions (e.g., \verb|^(/[^/ ]*)+/?$|) to identify all paths and equations in content highlighted with \textit{code}. For identifying software and terminologies, we collected all SO tags (e.g., ``mysql'', ``andriod'', ``django'', and ``hashtable'')\footnote{https://stackoverflow.com/tags} as our comprehensive dictionary. We then used fuzzy matching to identify such software and terminologies in the code dataset based on the dictionary. We used \textit{fuzzywuzzy} library\footnote{https://pypi.org/project/fuzzywuzzy/} for our implementation. The reason we selected SO tags as our dictionary is that they contain a diverse and wide range of software names and technologies~\cite{ye2016software}. we discarded those cases from our code dataset. 
    \item \textbf{Cleaning up text dataset.} Users sometimes highlight code data with text formatting tags. As it goes against the SO guidelines, we need to identify the code that is highlighted with text formatting tags (i.e., \textit{Bold, Italic, Heading}). To identify the code content in text dataset, we cross-checked the code dataset and text dataset and examined if any content was cross-highlighted in both code and text formatting. We removed such instances from text data. For instance, if we find ``BufferedReader.close()'' are both highlighted with Code and Bold formatting, we only keep the instances in code data and remove the instances from text data. 
    \item \textbf{Removing other HTML tags.} We used regular expressions to remove other HTML tags rather than the targeting tags. For example, for the \textit{Bold} model, we only kept the <b> and <strong> tags to identify the bold highlighting in sentences and removed other highlighting tags. 
\end{itemize}

Now we can construct our training and testing set. We divided the post into sentences and extracted the corresponding formatting tags described in Section~\ref{sec:data}. In other words, each data point is a highlighted instance (a sentence containing at least one type of information highlighting). When preparing the training data, we consider the content highlighted with different formatting types as the entities in the NER model. We annotate the tagged words with their formatting tags, considering their position in the phrase. For example, in the sample highlighted sentence shown in Figure~\ref{fig:NER_layers}, i.e., ``You will have to use \colorbox{lightgray}{client\_wait\_for()}.'', ``client\_wait\_for()'' is formatted with a code tag. To label this sentence, we assign the label ``B-code'' to the highlighted word, indicating that it is the beginning of a highlighted Code content, while the other words in the sentence are labeled as O (Outside). If multiple words are formatted, we used I (Interior) and E (Ending) to label the highlighted content's internal words and the ending word. For example, ``you need to run.\colorbox{lightgray}{git push -\phantom{}-force -\phantom{}-all} first'' contains more than one word in the highlighted Code content. So the labels for the phrase would be (B-code: ``git''), (I-code: ``push''), (I-code: ``-\phantom{}-force'') and (E-code: ``-\phantom{}-all''). Note that we train separate models for each highlighting type. When preparing the dataset for a particular highlighting type, if a sentence has other highlighting types rather than the one we target, we omit other types of highlighting types from the sentence.

\begin{table}[h]
    \centering
    \caption{Number of sentences in train and test datasets.} \label{tab:Datasetmodel}
    \begin{tabular}{|p{1.5in}|p{1.8in}|p{1.6in}|}
    \hline
        \textbf{Type} & \textbf{Training set} & \textbf{Testing set}\\ \hline
        Bold & 521308 & 140539 \\ \hline
        Italic & 477690 & 104128 \\ \hline
        Code  & 1410890 & 406960 \\ \hline
        Heading  & 131976 & 21028 \\ \hline
    \end{tabular}
\end{table}

Following prior studies, we partitioned the dataset into an 80:20 ratio for training and testing purposes~\cite{qiao2020deep,alrashedy2020scc++,yang2023does}. We trained our models in 3 epochs where the learning rate is 0.001. We employ two training settings.


We use 32 as the size of the batch in the fixed-size setting (i.e., \textit{fixed}). Models based on BERT architecture are trained in 0.05 learning rate using 16 batches(fixed) in 3 epochs.




\begin{table}[!ht]
    \centering
    \caption{Models parameters.}\label{ModelParameters}
    \begin{tabular}{|p{1.5in}|p{1.5in}|p{1.5in}|}
    \hline
        \textbf{Parameter} & \textbf{CNN-based} & \textbf{BERT-based}\\ \hline
        Epochs & 3 & 3\\ \hline
        Learning rate & 0.001 & 0.05\\ \hline
        Batch size & 32 & 16\\ \hline
        
    \end{tabular}
    \label{tab:parameters}
\end{table}


We ran our experiments on a Linux server with four Nvidia RTX 3090 GPUs, AMD Ryzen 48-Core CPU with 256 GB Ram. For simplicity, we call the models that are trained using the incremental size of batches \textit{incremental models} and those using the fixed size of batches \textit{fixed models}.

\vspace{0.1cm}
\noindent\underline{Evaluation:} We use precision, recall, and F1-score to measure the performance of models, following prior studies evaluating NER models~\cite{esuli2010evaluating,jiang2016evaluating}. Instead of using the exact match (i.e., a true positive is counted when all of the tokens of a named entity are labeled correctly, both the length and type), we decide to use a partial match~\cite{nadeau2007survey,tanabe2005genetag}, in which we consider if each individual token is predicted correctly. The partial match allows us to measure the performance of the segmentation aspects of highlighted content when the content has multiple words. More specifically, we calculate the precision and recall as follows:

$$Precision = \frac{Number\ of\ words\ that\ are\ correctly \ predicted}{Number\ of\ words\ that\ are\ predicted\ by\ the\ model}$$

$$Recall = \frac{Number\ of\ words\ that\ are\ correctly \ predicted}{Number\ of\ words\ that\ are\ supposed\ to\ tag}$$

For instance, suppose for a long highlighted sequence of words (e.g., \colorbox{lightgray}{sudo netstat -antp | fgrep LISTEN}), while only the words \colorbox{lightgray}{sudo netstat} and \colorbox{lightgray}{| fgrep LISTEN} are highlighted correctly, the words ``-antp |'' were missing. In the exact match, such a case is considered incorrectly tagged, while in the partial match, the words \colorbox{lightgray}{sudo netstat} and \colorbox{lightgray}{| fgrep LISTEN} are considered as correctly predicted.

\section{Results}\label{sec:results}
\subsection{\rqone{}}\label{sec:rq1}

\noindent\textbf{Results:}
\textbf{Overall, 47.6\% (14,845,929 out of 31,169,429) of the studied answers have information highlighted.} Among all the answers having information highlighted, an average of 10.6\% and a median of 7.1\% of the text (in words) is highlighted and each answer has an average of 3.9 and a median of 2 highlighting instances.

\begin{table*}
\centering
\caption{The answer post-wise and highlighted instance-wise statistics of different formatting types.}\label{tab:post_wise}
\scriptsize
\begin{tabular}{|p{0.4in}|p{0.55in}|p{0.8in}|p{1in}|p{0.6in}|p{0.89in}|}
  \hline
  \textbf{Type} & \textbf{\%Answers} & \textbf{\#Highlighted instance per answer (mean/median/max)} & \textbf{\%Highlighted words per answer (mean/median/max)} & \textbf{\%Instances} & \textbf{\#Words per instance (mean/median/max)} \\
  \hline
  Bold & 11.3\% & 2.0/1.0/241  & 9.0\%/4.1\%/100\% & 12.0\% & 2.9/1.0/436 \\
  Italic & 7.2\% & 1.9/1.0/354  & 5.9\%/2.3\%/100\% & 7.3\% & 2.9/1.0/477 \\
  Delete & 0.07\% & 1.2/1.0/36 & 18.1\%/10.6\%/100\%  & 0.04\% & 15.7/8.0/701 \\
  Code & 38.5\% & 3.7/2.0/490  & 8.7\%/6.3\%/100\%  & 78.9\% & 1.4/1.0/4,669 \\
  Heading & 1.6\% & 2.1/2.0/168  & 9.3\%/3.9\%/100\% & 1.7\% & 3.4/2.0/327 \\
  \hline
\end{tabular}
\end{table*}

\textbf{\emph{Code} is used the most frequently among all studied formatting followed by \emph{Bold} and \emph{Italic}.} Table~\ref{tab:post_wise} presents the statistics of different formatting types. We observe that \emph{Code} is the most frequently used type. 38.5\% of the studied answers have content highlighted with \emph{Code}. Moreover, 78.9\% of the highlighted instances are \emph{Code}. In addition, the studied answers have two \emph{Code} instances on the median, which is larger than other formatting types except \emph{Heading}. This finding is expected since users usually discuss programming problems on SO and it is common to highlight source code in the text. \emph{Bold} and \emph{Italic} are the most frequently used formats besides \emph{Code}. They are used in 11.3\% and 7.2\% of answers, respectively. \emph{Delete} is rarely used, only 0.07\% of the answers use it.

\textbf{In general, the length of highlighted content is short.} The median length of the content highlighted with \emph{Code}, \emph{Bold}, \emph{Italic}, \emph{Deleting}, and \emph{Heading} are 1, 1, 1, 8, and 2 words, respectively. Users tend to highlight single word or phrase using \emph{Code}, \emph{Bold}, and \emph{Italic}.  Compared with other formatting types, \emph{Delete} instances are longer. 

\rqbox{Information highlighting is prevalent on SO, i.e., 47.6\% of the answers use the studied formatting to highlight information. 38.5\% of the answers use \emph{Code}, which is the most frequently used format, followed by \emph{Bold} (11.3\%) and \emph{Italic} (7.2\%). In general, the length of highlighted content is short.}

\subsection{\rqtwo{}}\label{sec:rq2}

\noindent\textbf{Results:}
\textbf{\emph{Code} is mainly used to highlight source code elements, such as identifiers (63.5\%), programming language keywords (9.9\%), and statements (7.0\%).} Table~\ref{tab:codelabel} presents the derived types of content that is highlighted with \emph{Code} and the distribution of each type. \emph{Code} is used the most frequently to highlight identifiers in source code, such as the name of classes, methods, parameters, and variables. We also observe that 59\% of the highlighted identifiers appearing in the code block of their corresponding answers. That is said, \emph{Code} is commonly used to highlight identifiers in text for referring to the corresponding identifiers in the code blocks. This is reasonable since users usually need to refer to a unit in the source code when discussing the solution. In addition, 9.9\% and 7.0\% of the studied instances are for Keyword, and Statement, respectively.

\textbf{\emph{Code} is also used to highlight content other than source code, such as Software (4.9\%), Terminology (1.8\%), Equation (5.2\%), and Version (0.5\%).} From Table~\ref{tab:codelabel}, we observe that users also use \emph{Code} to highlight content other than code. For instance, in 5.2\% of the cases, \emph{Code} is used to highlight content related to equations. For example, in an answer, answerer discuss the time complexity of binary search ``This is \colorbox{lightgray}{O(log n)}'', which is highlighted using \emph{Code}\footnote{https://stackoverflow.com/questions/17117375}. \emph{Code} sometimes is used to emphasize the names of a software, a framework, and a lib. For instance, the answerer of an answer mentioned ``You are using \colorbox{lightgray}{Mysql} as DB ...'' and show a piece of code to demonstrate the database connection between Java and Mysql\footnote{https://stackoverflow.com/questions/24586043}.

\begin{table}[!ht]
    \centering
    \caption{The definition and distribution of types of content highlighted with \emph{Code} formatting.}\label{tab:codelabel}
    \begin{tabular}{|p{0.9in}|p{3.4in}|p{0.6in}|}
    \hline
        \textbf{Type} & \textbf{Definition (example)} & \textbf{Count (\%)}\\ \hline
        Identifier & The identifier in source code, e.g, ``ArrayList''. &  245 (63.6\%) \\ \hline
        Keyword & The keywords in programming languages, e.g., ``for'' and ``public''. &  38 (9.9\%) \\ \hline
        Statement & A statement of source code, e.g., ``plot = last.plot()''. & 27 (7.0\%) \\ \hline
         Equation & Mathematical equation, operator, or number, e.g., ``O(log n)''. & 21 (5.2\%) \\ \hline
        Software & The name of softwares, frameworks, tools, and libs, e.g., ``Tensorflow'' and ``MySQL''. & 19 (4.9\%) \\ \hline
        Path & Path and files name, e.g., ``src/main/java''.  & 17 (4.4\%) \\ \hline
       
        Terminology & Terminologies related to programming, e.g., ``hash table''. & 7 (1.8\%) \\ \hline
        Cmd & Command, e.g., ``cloud-init init'' & 5 (1.3\%) \\ \hline
        Version & Version information of a software, e.g., ``62.1''. & 2 (0.5\%) \\ \hline
        Other & Other than the above defined types. & 4 (1\%) \\ \hline
    \end{tabular}
\end{table}

\rqbox{Although \emph{Code} is mainly used to highlighted the content related to source code, it is also used to highlight content other than source code, such as Software (4.9\%), Terminology (1.8\%), Equation (5.2\%), and Version (0.5\%).}

\subsection{\rqthree{}}\label{sec:rq3}

\begin{table*}[!ht]
    \centering
    \caption{The definition and distribution of different types of content highlighted with \emph{Bold, Italic, Heading, Delete}.}\label{tab:textlabel}
    \scriptsize
      \resizebox{\textwidth}{!}{%
    \begin{tabular}{|l|p{2.1in}|l|l|l|l|}
    \hline
        \textbf{Type} & \textbf{Definition (example)} & \textbf{Bold} & \textbf{Italic} & \textbf{Delete} & \textbf{Heading} \\ \hline
        Update & The update and new edit in the post, e.g., ``\textbf{Update:} According to your comments, I think you will want to try gzfile() in read.table()''. & 22 (12.4\%) & 2 (1.2\%) & 0 & 0 \\ \hline
        Equation & Same as the definition of Equation in Table~\ref{tab:codelabel}. & 3 (1.7\%) & 2 (1.2\%) & 0 & 0 \\ \hline
        Source code & The union of Identifier/Keyword/Statement/Operator/Cmd/Path in Table~\ref{tab:codelabel}, e.g., ``I'd suggest the usage of \textbf{startupOrder} to configure the startup of route though''. & 51 (28.8\%) & 52 (30.8\%) & 0 & 1 (2.8\%) \\ \hline
        Caveat & A reminder or warn of in which context or condition the provided solution works or does not work, e.g., ``\textbf{It won't work in that shell though, you need to open a new one}''. & 33 (18.6\%)  & 43 (25.4\%) & 0 & 1 (2.8\%) \\ \hline
        Reference & Reference to internal or external resource, e.g., ``<a href="http://api.jquery.com/not/">.not()</a>''. & 12 (6.8\%) & 17 (10.1\%) & 0 & 0 \\ \hline
        Results & Results or output.  & 3 (1.7\%) & 2 (1.2\%) & 0 & 0 \\ \hline
        Terminology & Same as the definition of Terminology in Table~\ref{tab:codelabel}. & 12 (6.8\%) & 21 (12.4\%) & 0 & 0 \\ \hline

        Heading & The title of a section, e.g., ``\textbf{WORKING EXAMPLE}''. & 27 (15.3\%) & 2 (1.2\%) & 0 & 33 (91.6\%) \\ \hline
        Options & Options in software, e.g., ``you have to enable Less \textbf{Secure Sign-In} in your google account''. & 4 (2.3\%) & 1 (0.6\%) & 0 & 0 \\ \hline
        Extent & A word/phrase to express extent, e.g., ``at least with Hibernate it assumes you want to use a global "hibernate" sequence for \emph{all} tables, which is just stupid.'' & 4 (2.3\%) & 14 (8.3\%) & 0 & 0 \\ \hline
        Version & Same as the definition of Version in Table~\ref{tab:codelabel}. & 0 & 3 (1.8\%) & 0 & 0 \\ \hline
        Delete & Deleted outdated/wrong description, e.g., ``Just use \sout{KVO} KVC''. & 0 & 0 & 3 (100\%) & 0 \\ \hline
        Other & Other than the above defined types. & 6 (3.4\%) & 10 (5.9\%) & 0 & 1 \\ \hline
    \end{tabular}
    }
\end{table*}

\noindent\textbf{Results:}
\textbf{Both \emph{Bold} and \emph{Italic} formatting are most frequently used to highlight content related to source code.} Table~\ref{tab:textlabel} presents the distribution of each \emph{Text} type. We observe for certain types, \emph{Bold} and \emph{Italic} share similar patterns. For instance, in 28.8\% of the \emph{Bold} instances and 30.8\% of the \emph{Italic} instances, content related source code (i.e., Source code) is highlighted. That is said, users also frequently use \emph{Bold} and \emph{Italic} to highlight source code other than using \emph{Code} formatting. Users probably use \emph{Bold}, \emph{Italic}, and \emph{Code} exchangeably, although SO suggests users to tag inline code using \emph{Code} format~\cite{StackExchangeMarkdown}. Interestingly, we also observe in some cases, the community members changed formatting of the code-related content from \emph{Bold} and \emph{Italic} to \emph{Code}. For instance, in an answer, a user change the formatting for a funciton ``strncmp()'' from \emph{Italic} to \emph{Code}\footnote{https://stackoverflow.com/posts/18437465/revisions}.

\textbf{Users frequently use both \emph{Bold} and \emph{Italic} to highlight Caveat, Reference, and Terminology.} In some cases, it is important to mention the condition or context in which a provided solution works (i.e., Caveat). We notice that such information is usually highlighted in answers. For instance, in a SO answer, the answerer reminded readers ``Donot forget add jmtp.dll files (that comes up with jmtp download) as a native library for more info see my answer on ...'' using \emph{Bold}.\footnote{https://stackoverflow.com/questions/6498179}. \emph{Bold} and \emph{Italic} are used to highlight Reference and Terminology, although \emph{Italic} is used more often than \emph{Bold}.

\textbf{Users tend to highlight the content of types Update and Heading with \emph{Bold}, while \emph{Italic} is more likely to be used for the type of Extent.} From Table~\ref{tab:textlabel}, we can see differences between \emph{Bold} and \emph{Italic}. For instance, \emph{Bold} is more frequently used to highlight Update (12.4\%) and Heading (15.3\%) compare with \emph{Italic}. Interestingly, users also use \emph{Bold} to highlight a heading. One possible reason is that the rendering effect for \emph{Bold} and \emph{Heading} are similar. Prior studies reveal that answers on SO are easy to become obsolete~\cite{ragkhitwetsagul2019toxic,zhang2019empirical}. Therefore, it is not surprising to observe that users highlight the update using \emph{Text} formatting. Different from \emph{Bold}, \emph{Italic} is used more frequently to highlight a word/phrase to express the extent (i.e., Extent).

For \emph{Delete}, it is all used to highlight the obsolete or incorrect content. In terms of \emph{Heading}, we observe that in 91.6\% of the instances, users used it to highlight heading, which is expected. However, we also observe in one case for highlighting Source Code and one case for emphasizing Caveat.

\rqbox{Apart from source code, both \emph{Italic} and \emph{Bold} are used frequently to highlight content in types of Caveat, Reference, Terminology, and Update.}

\subsection{\rqfour{}}\label{sec:rq4}

\vspace{0.1cm}

\textbf{Out of the four trained models, the Code model obtained the highest F1 score of 0.69 (CNN).} Table~\ref{tab:EvaluationModel} presents the results of our trained models in terms of precision, recall, and F1 using CNN and BERT sizes of batches. In general, in terms of F1, we observe that the Code models (F1 score is 0.162 and 0.69 using BERT and CNN models) perform better than other models. This is reasonable, since first for code formatting, we have much more training data than other types. Secondly, code typically has certain special characters (e.g., ``.'', and ``()'' in API ``class.function()'') and keywords (e.g., ``foreach''), making it easy for the model to identify. Our observation indicates that it is easier to build a model to identify code and highlight them on SO posts than other types of formatting. 

We also notice that for code data, CNN-based models outperform BERT-based models by a large margin, typically in terms of recall. The CNN-based model for code achieves a recall of 0.66, while the BERT-based model only has a recall of 0.093. In other words, the CNN-based model recognizes and covers more real code-related content than the BERT-based model. One possible explanation is that the CNN-based model is smaller than BERT and makes it easier to train and learn the patterns to recognize the code-related content to highlight. Another possibility is that BERT is pre-trained on natural language and not for code, which makes it hard to recognize code-related content.

\begin{table*}[htp]
    \centering
    \caption{The performance of models on all formatting types \emph{Bold, Italic, Heading, Code} with different training settings. \textit{CNN} denotes the CNN-based models and \textit{BERT} denotes the BERT-based models. }\label{tab:EvaluationModel}
    \resizebox{\textwidth}{!}{%
    \begin{tabular}{|p{0.9in}|p{0.7in}|p{0.7in}|p{0.6in}|p{0.6in}|p{0.55in}|p{0.6in}|}
    \hline 
        \textbf{Type} & \textbf{Precision (\textit{CNN})}& \textbf{Precision (\textit{BERT})} & \textbf{Recall (\textit{CNN})}& \textbf{Recall (\textit{BERT})} & \textbf{F1 (\textit{CNN})}& \textbf{F1 (\textit{BERT})}\\ \hline
        $Model_{Bold}$ & 0.57 & 0.63 & 0.05 & 0.006 & 0.09 & 0.011 \\ \hline
        $Model_{Italic}$ & 0.64 & 0.53 & 0.01 & 0.0025 & 0.02 & 0.005 \\ \hline
        $Model_{Heading}$ & 0.50  & 0.49 & 0.04 & 0.0073
 & 0.08 & 0.0143\\ \hline   
        $Model_{Code}$ & 0.72 & 0.65 & 0.66 & 0.093 & 0.69 & 0.162\\ \hline

    \end{tabular}
    }
\end{table*}

\textbf{Our CNN-based models achieve an acceptable precision ranging from 0.50 to 0.72 for different formatting types while obtaining low recall.}  Overall, except for the Code models, other models have higher precision and lower recall. For models for text formatting, Bold, and Italic models can achieve at least a precision of 0.57 from the models. While models for text formatting struggle from low recall. The results suggest that as long as the text formatting models recommend the content to highlight, the accuracy is acceptable, while the models \textbf{miss} most of the content that is supposed to be highlighted (low recall). One possible reason is the training data for those three types are much smaller than code and the patterns for text formatting types are more diverse, which makes it challenging to learn models for text formatting. 
\begin{table*}[!ht]
    \centering
    \caption{Successful examples of content highlighted by our models (using the incremental setting) along with original text for different types of formatting. Note that since our models are trained for one specific type of formatting, therefore we only show the content that is highlighted by the specific model in the examples. }\label{tab:ExampleSentences}
    \resizebox{\textwidth}{!}{%
    \begin{tabular}{|p{1.3cm}|p{6cm}|p{6cm}|}
    \hline
       \textbf{Type} & \textbf{Original text} &  \textbf{Text highlighted with our models (Incremental).}\\ \hline
        
        Bold & \textbf{NOTE : THIS MIGHT ALREADY SOMETHING THAT YOU HAVE TRIED.} This is probably due to the fact that you are using \colorbox{lightgray}{--trusted-host=pypi.python.org}. & \textbf{NOTE : THIS MIGHT ALREADY SOMETHING THAT YOU HAVE TRIED}. This is probably due to the fact that you are using --trusted-host=pypi.python.org.\\ \hline

         Code & For more info on searching for the elements on DOM such as \colorbox{lightgray}{getElementById},\colorbox{lightgray}{querySelector}, please refer here. & For more info on searching for the elements on DOM such as \colorbox{lightgray}{getElementById},\colorbox{lightgray}{querySelector}, please refer here.\\ \hline     
        
        Italic & Side note: Your code is falling prey to what I call \textit{The Horror of Implicit Globals} because you never declare filename. & Side note: Your code is falling prey to what I call \textit{The Horror of Implicit Globals} because you never declare filename. \\ \hline
       
        Heading & {\large OPTION 1: GIT STASH}
        \par You may postpone all current changes of your files in the git stash cache. After that the files are equal to former checkout state. & {\large OPTION 1: GIT STASH}
        \par You may postpone all current changes of your files in the git stash cache. After that the files are equal to former checkout state. \\ \hline

    \end{tabular}%
    }
\end{table*}

In Table~\ref{tab:ExampleSentences}, we present several successful examples, in which the corresponding content is highlighted with correct formatting types. In the Code example, the model identifies all code-related content in the post correctly, such as \colorbox{lightgray}{getElementById} and \colorbox{lightgray}{querySelector}. For Italic, the model can identify ``The Horror of Implicit Globals'' in the sentence. For Heading, the model successfully identifies the first heading sentence.

\vspace{0.1cm}
\noindent\underline{Case Study of Failure Cases:} We also investigate the failure cases of the models to provide some insights for future research. The failure cases could be categorized into the following three families:
\begin{itemize}
    \item \textbf{Misidentification:} Although the content that needs highlighting is successfully identified, incorrect formatting types are recommended. In other words, the content is not identified by the correct model.
    \item \textbf{False Identification:} The model identifies the content that is not supposed to be highlighted. 
    \item \textbf{Missing Identification:} The model misses the content that is supposed to be highlighted.
\end{itemize}

We calculate the distribution of three failure cases over each model using the confusion matrix. For instance, to compute the number of missing identification cases from $Model_{Bold}$, we count the number of false negative instances in the confusion matrix on the testing data generated by $Model_{Bold}$.

\begin{table}[h]
\centering
\caption{The distribution of different failure types among CNN models of different formatting types.}
\label{tab:disfailure}
\footnotesize
\resizebox{\textwidth}{!}{%
\begin{tabular}{|p{3.3cm}|p{1.5cm}|p{1.5cm}|p{1.5cm}|p{1.5cm}|p{1.5cm}|}
\hline
\multicolumn{1}{|p{0.5cm}|}{\multirow{2}{*}{\textbf{Failure Type}}} & \multicolumn{4}{p{5cm}|}{\textbf{Formatting Type}} \\\cline{2-5}
\multicolumn{1}{|p{1cm}|}{} & Bold & Code & Italic   & Heading  \\ \hline
{Misidentification}  & 0.86\% & 4.09\% & 2.65\% &  15.19\%\\ \hline
{Missing Identification} & 95.6\%\ & 52.77\% & 96.67\% &  81.1\% \\ \hline
{False Identification}  & 3.55\% & 43.13\% & 0.67\% &  3.63\%  \\ \hline
\end{tabular}%
}
\end{table}

\begin{figure}
    \subfloat[]{\includegraphics[width=.52\textwidth]{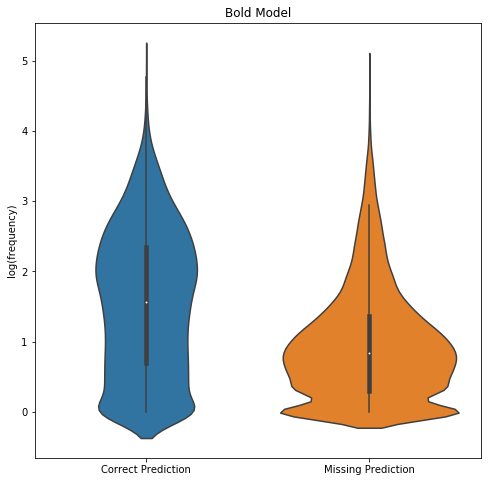}}\hfill
    \subfloat[]{\includegraphics[width=.52\textwidth]{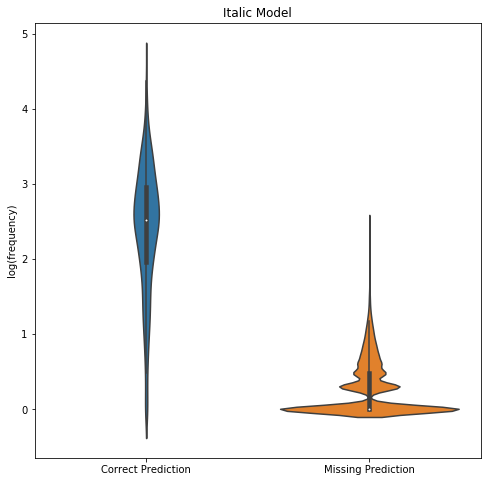}}
    \\[\smallskipamount]
    \subfloat[]{\includegraphics[width=.52\textwidth]{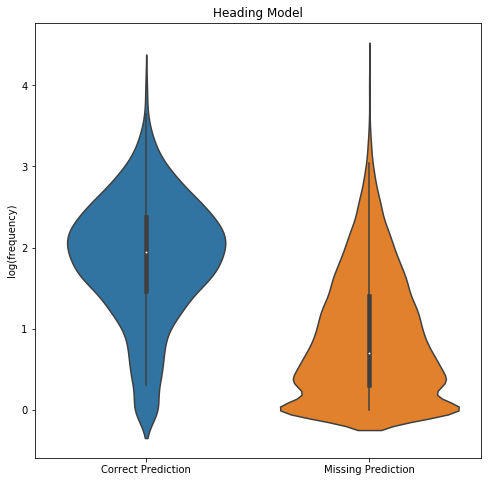}}\hfill
    \subfloat[]{\includegraphics[width=.52\textwidth]{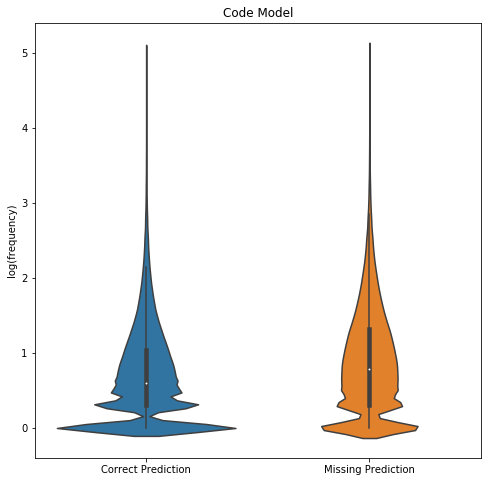}}
    
    \caption{Comparison of the frequency distribution of words that are not identified (missing identification) and those that are correctly predicted (correct prediction) on different models.}\label{fig:WordFrequency}
\end{figure}

Table~\ref{tab:disfailure} presents the distribution of those three types of failures across different models. Overall, the majority of the failure cases are missing identification  All models suffer from high missing identification except Code models, which explains the low recall in Table~\ref{tab:EvaluationModel}, typically for text formatting tag (Bold, Italic, and Heading) models. One possible explanation is that the models are easy to learn the frequent highlighted words while struggling to learn less frequent words (i.g., long-tail knowledge)~\cite{kandpal2022large,mireshghallah2022memorization}. 
To test this assumption, we count the frequency of the words that are correctly predicted and the words that models in the training data do not identify. 
We examine the recommendation results produced by each CNN-based model.
Table~\ref{tab:CorrectMissingFrequencies} presents the mean/median frequency of words that are correctly predicted and those that are not identified by models. On top of it, Figure~\ref{fig:WordFrequency} compares their distributions. We observe that the mean and median frequency of words that are predicted correctly is much larger than those of words that are not identified in all models, except the median value of Code. 

\begin{table*}[htp]
    \centering
    \caption{The mean/median frequency of words that are correctly predicted (correct prediction) and those that are not identified by models (missing identification) in different models.}\label{tab:CorrectMissingFrequencies}
    \resizebox{\textwidth}{!}{%
    \begin{tabular}{|p{0.8in}|p{2.0in}|p{2.0in}|}
    \hline 
        \textbf{Type} & \textbf{Mean/Median of missing identification} & \textbf{Mean/Median of correct prediction} \\ \hline
        Bold & 127.7/7 & 441.5/37.0 \\ \hline
        Italic & 2.39/2.52 & 959.9/332.0 \\ \hline
        Heading & 1.86/1.94 & 241.24/89.0\\ \hline
        Code 50\% & 0.699/0.602 & 1931.04/8.0 \\ \hline

    \end{tabular}
    }
\end{table*}

We also observe a small portion of misidentification cases and False identification. For instance, in example 1 in Table~\ref{tab:FailureCases}, the original text is highlighted with Heading, while our model identifies the content but formats it using bold. One possible explanation is that those types of formatting share common usage patterns to some extent and confuse the models. We actually observed such a phenomenon in Section~\ref{sec:rq3}. Users sometimes highlight the same content with both Bold Italic tags. 
For instance, there are many cases in which if there is any update on the posts, the keyword ``update’’ will be highlighted by different types of formatting, i.e.,  ``update'' was highlighted the most by Bold tag (0.028\%) followed by other tags (Heading 0.0054\%, Code 0.00269\% and Italic 0.00249\%). Therefore, when recommending the information to highlight, if multiple models recommend the same content (e.g., both the Bold model and Italic model recommend ``update’’ to be highlighted), we use the formatting tag that is returned by the model with the highest confidence as the final tag. This decision is based on the assumption that the recommendation from the model with the highest confidence aligns with the predominant choice of the community. By prioritizing the recommendation from the most confident model, we aim to ensure consistency and accuracy in the formatting of highlighted information, reflecting the collective preferences of the user community. Another option is that we can return all recommendation results with confidence scores returned by each model and allow users to select, like Google Query Suggestion~\cite{googlequerysuggestion}.

\begin{table*}[!ht]
    \centering
    \caption{Failure cases predicted by CNN-based models for all the types \emph{Bold, Italic, Heading, and code.}}\label{tab:FailureCases}
    \resizebox{\textwidth}{!}{%
    \begin{tabular}{|p{1.5cm}|p{1.5cm}|p{6cm}|p{6cm}|p{6cm}|}
    \hline
       \textbf{ID} & \textbf{Type} & \textbf{Original text} &  \textbf{Text highlighted with our models (Incremental)}. & \textbf{Failure type}\\ \hline
       1 & Bold & {\large Swift 5.2.x }\par First of all, you need to declare an "easy to use" typealias for your block: & \textbf{Swift 5.2.x} \par
        First of all, you need to declare an "easy to use" typealias for your block: & \textbf{Misidentification:} Detected heading as bold. \\ \hline
       2 &  Bold & If you want to extend it's expiration date, IT WILL THROW THE EXCEPTION! & If you want to extend it's expiration date, IT\textbf{ WILL THROW THE EXCEPTION!} & \textbf{False Identification:} detected normal texts as bold. \\ \hline
       3 & Bold & \textbf{From now, your frontend application will use access token in the Authorization header for every request.} & From now, your frontend application will use access token in the Authorization header for every request. &  \textbf{Missing Identification}: missed Bold.\\ \hline

       4 & Italic &  Disclaimer: I currently work for Mapbox. &  Disclaimer: \textit{I currently} work for Mapbox. & \textbf{False Identification:} detected normal texts as Italic.\\ \hline
       
        5 & Heading & Close Sublime and Quit It. Open up your desired project that you want to use Sublime with in Unity & {\large Close Sublime} and Quit It. Open up your desired project that you want to use Sublime with in Unity & \textbf{False Identification:} detected normal texts as Heading. \\ \hline
        
        6 & Heading & {\large Because there are no spaces in the text.}
Chinese is commonly written without any space characters between words or sentences. The ASS subtitle format, having been originally designed by and for European language speakers, unfortunately only breaks words on actual space characters. Your subtitle file does not contain any space characters, so the lines are never broken.
        & Because there are no spaces in the text.
Chinese is commonly written without any space characters between words or sentences. The ASS subtitle format, having been originally designed by and for European language speakers, unfortunately only breaks words on actual space characters. Your subtitle file does not contain any space characters, so the lines are never broken. & \textbf{Missing Identification}: Missing Heading for ``Because there are no spaces in the text''.
\\ \hline
       7 &   Code & setTimeout seems like the right guy for the job. 
        To \textit{off} a click use jQuery's \colorbox{lightgray}{.off()} Method. Also, the \colorbox{lightgray}{.one()} method to attach a callback only \textit{once} : & \colorbox{lightgray}{setTimeout} seems like the right guy for the job.
        To off a click use jQuery's \colorbox{lightgray}{.off()} Method. Also, the \colorbox{lightgray}{.one()} method to attach a callback only once : & \textbf{False Identification:} Detected normal text as code.\\ \hline
      8 &   Code & Most likely that the website you are trying to scrape is a dynamic website, meaning that it is Javascript generated code and can't be scraped with only \colorbox{lightgray}{requests} and \colorbox{lightgray}{beautifulsoup} . & Most likely that the website you are trying to scrape is a dynamic website, meaning that it is Javascript generated code and can't be scraped with only requests and beautifulsoup .& \textbf{Missing Identification}: Missed the Code tag.\\ \hline
    \end{tabular}}
\end{table*}

\rqbox{Our models achieve a precision ranging from 0.50 to 0.72 for different formatting types. It is easier to build a model to recommend Code than other types. Models for text formatting types (i.e., Heading, Bold, and Italic) suffer low recall. Our analysis of failure cases indicates that the majority of the failure cases are due to missing identification. One explanation is that the models are easy to learn the frequent highlighted words while struggling to learn less frequent words (i.g., long-tail knowledge).}

\section{Discussion}\label{sec:disc}
\subsection{Usefulness of recommendation models}
\begin{figure}
    \centering
    \includegraphics[width=\linewidth]{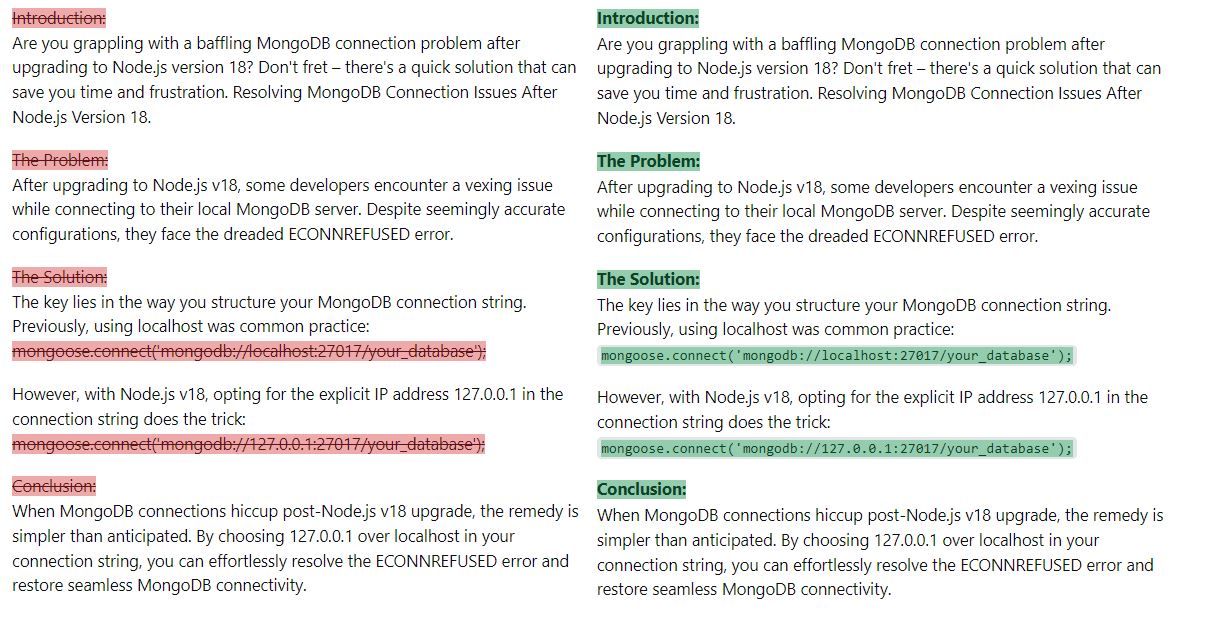}
    \caption{An example of a post where our formatting improvements were approved by the community.}
    \label{fig:example_usefulness}
\end{figure}

SO encourages the community to improve the quality of posts by editing them and proper edits will be approved by trusted community members\cite{SOEdit}. To demonstrate the usefulness of the proposed recommendation models, we randomly sampled 20 new SO posts that are outside of our dataset, and applied our CNN-based model on them to recommend the information to highlight. We found that in 9 posts, the content recommended by our models was already highlighted. For the rest 11 posts, we highlight the content recommended by the model. All our recommended edits in those 11 posts were accepted by the community. For instance, in Figure~\ref{fig:example_usefulness}, our model recommends highlighting the subsection title (e.g., Introduction and The Problem) in \textit{Bold} and highlighting the code with \textit{Code}. We posted the edits and all of our edits were approved by the community\footnote{https://stackoverflow.com/posts/76886399/revisions}.

\subsection{Implication of our findings}
\textbf{Future search should consider the highlighted content for the downstream tasks that leverage information from the SO answers.} In RQ3, we observe that users tend to highlight important information for a provided answer using \emph{Text} formatting, such as a reminder or warn of in which context or condition the provided solution works or does not work (i.e., Caveat), an update and new edit in answers (i.e., Update). Such content is informative and important to users when learning and applying knowledge from the answers. Therefore, it is substantial to consider such information for the downstream tasks that leverage information extracted from SO answers, such as answer summarization~\cite{ren2019discovering}, and API documentation enrichment~\cite{li2018improving,treude2016augmenting}.

\textbf{Our findings provide insights to improve future research in developing approaches for automatic information highlight.} In RQ4, we investigated the potential of deep learning NER models in recommending information to highlight on SO posts. Code is easier to identify than other types of formatting. Furthermore, all text-related models (i.e., Italic, Bold, and Heading) suffer from low recall. We reveal one reason is long-tail knowledge. The model tends to memorize frequently highlighted words and miss the recognition of the less-frequent (e.g., tail knowledge). Future research is encouraged to leverage techniques~\cite{zhang2021deep}, e.g., data augmentation and class rebalancing for tail knowledge.


\subsection{Threats to validity}

\noindent\textbf{Internal Validity:} Our study involved qualitative analysis in RQs. To reduce the bias, each instance was labeled by two of the authors, and discrepancies were discussed until a consensus was reached. We also showed that the level of inter-rater agreement of the qualitative studies is high. In RQ4, we evaluate our models using precision, recall, and F-1 score with partial match. Although, there might be other metrics that could be used for our task, those metrics are commonly used in NER models, and we follow previous studies~\cite{esuli2010evaluating,jiang2016evaluating}. In RQ4, we investigated the potential of a CNN-based model and trained it from scratch based on our own dataset. We also investigated a pre-trained model BERT and fine-tuned it for our task. However, our finding might not be generalized to other models. We encourage future research to investigate more models for the task.

\vspace{0.1cm}
\noindent\textbf{External Validity:} One external threat is that it is not clear whether our findings still hold on other Q\&A websites. We needed to conduct several qualitative analysis in our RQs; however, it is impossible to manually study all instances. To minimize the bias when conducting our qualitative analysis, we took statistically representative random samples of all relevant revisions, in order to ensure a 95\% confidence level and 5\% confidence interval for our observations.

\noindent\textbf{Content Validity:} 
Content validity refers to the extent to which a measure represents all facets of a given construct. In this study, we focus on \emph{Bold}, \emph{Italic}, \emph{Heading}, \emph{Delete}, and \emph{Code}. Stack Overflow also provides other types of formatting to highlight information, such as lists and links. To mitigate the threat, we focus on the most commonly used ones. Future research is encouraged to study more types of formatting.

\noindent\textbf{Construct Validity:} One threat to the construct validity is related to how we collected instances for training recommendation models for different formatting types. 
Although we made diligent efforts to clean up the dataset and eliminate inconsistencies from both code and text data, as discussed in Section~\ref{sec:rq4approach}, it is important to acknowledge that it may not have been possible to remove all inconsistencies. This could potentially pose a threat to the validity of our results.
However, our failure cases analysis in Section~\ref{sec:rq4} reveals that the proportion of misidentification is low, consistently less than 15.17\% across all models. This suggests that following the data cleaning process, the models are not easily confused among different types of formatting. This provides evidence regarding the effectiveness of our data cleaning efforts and increases the reliability of our results.
Another threat to the construct validity relates to the suitability of our evaluation metrics: Precision, Recall, and F1-measure based on partial match. Other metrics could be used for evaluating the recommendation system. We selected those metrics because these metrics have been used in many NER studies~\cite{esuli2010evaluating,jiang2016evaluating}. This threat is limited by the research. 
\section{Related work}\label{sec:related}

\subsection{Information highlighting}
Previous research has explored the benefits of highlighting information in various domains over the last decades~\cite{nguyen2015combining,ramirez2019understanding,strobelt2015guidelines,wilson2016crowdsourcing,wu2003improving}. Wu and Yuan have shown that highlighting could reduce the cognitive load and thus reduce reading time~\cite{wu2003improving}. Jorge et al. investigated the impact of highlighting text in the text classification tasks in the machine learning area and found that highlighting is effective in  reducing classification effort for humans~\cite {ramirez2019understanding}. Similarly, Nguyen et al. developed a tool to explain the output ML models by highlighting the portion of text and demonstrated its effectiveness in model explanation~\cite{nguyen2015combining}. 
Various studies have been done to investigate the impact of information highlighting in software engineering, such as code comprehension~\cite{beelders2016syntax,hannebauer2018does,palma2022fly,sarkar2015impact}, source code evolution~\cite{escobar2022spike}. For instance, Advait Sarkar investigated the effect of syntax highlighting on program comprehension and its interaction with programming experience~\cite{sarkar2015impact}, and found that syntax highlighting significantly improves task completion time. 
Different from prior studies that focus on understanding the influence of information highlighting, we focus on understanding the practice of information highlighting on SO.

\subsection{Knowledge retrieval on Stack Overflow}
Several studies have investigated developers' challenges and solutions in retrieving knowledge from technical Q\&A sites and provided some tools to facilitate the process~\cite{gottipati2011finding,nadi2020essential,xu2017answerbot,zhang2021comments}. For instance, Gottipati et al. found that in software forums it is a painstaking process for users to manually search through answers in various threads of posts~\cite{gottipati2011finding}, and they developed a customized search engine to find relevant answers from various software forums. Similarly, Xu et al. developed a technique called AnswerBot, which takes input as the technical question and generates an answer summary for the question~\cite{xu2017answerbot}. Zhang et al. pointed out an issue of current comment ranking and display mechanism on SO (e.g., a large portion of useful comments are not visible to users) and proposed an alternative ranking mechanism to alleviate the issue~\cite{zhang2021comments}. Sarah and Treude investigated the possibility of using two existing techniques, wordpartten and lexrank, to identify the essential sentences from a long SO answer~\cite{nadi2020essential}. Different from prior studies that focus on retrieving relevant or important answers from SO, we investigate what and how information is highlighted on SO. Our study could provide insights to enhance existing techniques.

\subsection{Recommendation system for Stack Overflow}

Previous studies developed different recommendation tools to ease the use of the Q\&A sites, such as recommending hyperlinks~\cite{Li2016discussion,Li2018LinkLiveDW}, tags~\cite{He2022PTM4TagST,maity2019,rekha2014,wang2019SOTagRec,wang2018entagrec++}, and similar questions~\cite{Wang2019IEA,yin2019}. Li et al. developed LinkLiv to recommend hyperlinks~\cite{Li2018LinkLiveDW}, which uses multiple features, including hyperlink co-occurrences in Q\&A discussions, and locations (e.g., question, answer, or comment).
Wang et al. developed EnTagRec, which combines two complementary lines in the statistics community Bayesian and frequentist~\cite{wang2018entagrec++}. Wang et al. developed SOTagRec combining a convolutional neural network and collaborative filtering model to infer tags for new postings by studying the historical postings and their tags~\cite{wang2019SOTagRec}. Maity et al. developed DeepTagRec~\cite{maity2019}, which learns from the content representation of question title and body, and recommends appropriate question tags on Stack Overflow.
Wang et al. developed an approach that combines topical interest, topical expertise, and activeness to recommend answers for new questions~\cite{Wang2019IEA}.
Similar to these studies, we investigate the potential of developing models for recommending information to highlight in SO answers. We adapted Named Entity Recognition (NER) models to identify the content that needs to be highlighted.

\section{Conclusion}\label{sec:conclusion}
This paper is the first large-scale study of information highlighting in the text description of SO answers. We find that information highlighting is prevalent, with 47.6\% of the 31,169,42 studied answers having text highlighted. We propose a terminology to categorize highlighted text in SO answers and find that source code content (e.g., identifiers and programming keywords) are frequently highlighted using \emph{Code} and other highlighting formats (i.e., \emph{Bold} and \emph{Italic}). Besides code, users also tend to highlight updates (e.g., updates of answers), caveats (i.e., a reminder or warning of in which context or condition the provided solution works or does not work), and references.
Further studies could put more effort into investigating how to use the highlighted content for downstream tasks (e.g., answer summarizing) that leverage information from SO answers, and provide tools that can suggest highlighted text for SO users. In addition, we also investigate the potential of recommending the content to be highlighted automatically by adopting named entity recognition (NER) models. 
Our experimental results highlight that our CNN-based models achieve precision scores between 0.5 and 0.72 across different formatting types, although they exhibit lower recall rates. On the other hand, while BERT demonstrates superior precision, it struggles with even lower recall. It's worth noting that CNN-based Code model performs exceptionally well, with an impressive F1 score of 0.71, outperforming other models in our study.
Although BERT usually has better performance in other tasks, in our case the CNN-based model worked better. In future research, we plan to investigate other more advanced models (e.g., LLaMA) and work on improving the recall rates for both models.

\bibliographystyle{natbib}
\bibliography{main}

\begin{thebibliography}{}

\bibitem[goo(2024)goo]{googlequerysuggestion}
 (2024).
\newblock {Google query suggestion}.
\newblock \url{https://www.google.com/support/enterprise/static/gsa/docs/admin/current/gsa_doc_set/xml_reference/query_suggestion.html}.
\newblock Accessed: 2023-04-10.

\bibitem[Ahmed {\em et~al.}(2022)Ahmed, Wang, Zhang, Chen, and Tian]{ahmed2022first}
Ahmed, S.~S., Wang, S., Zhang, H., Chen, T.-H., and Tian, Y. (2022).
\newblock A first look at information highlighting in stack overflow answers.
\newblock In {\em 2022 IEEE International Conference on Software Maintenance and Evolution (ICSME)\/}, pages 369--373. IEEE.

\bibitem[Alrashedy {\em et~al.}(2020)Alrashedy, Dharmaretnam, German, Srinivasan, and Gulliver]{alrashedy2020scc++}
Alrashedy, K., Dharmaretnam, D., German, D.~M., Srinivasan, V., and Gulliver, T.~A. (2020).
\newblock Scc++: Predicting the programming language of questions and snippets of stack overflow.
\newblock {\em Journal of Systems and Software\/}, {\bf 162}, 110505.

\bibitem[Beelders and du~Plessis(2016)Beelders and du~Plessis]{beelders2016syntax}
Beelders, T.~R. and du~Plessis, J.-P.~L. (2016).
\newblock Syntax highlighting as an influencing factor when reading and comprehending source code.
\newblock {\em Journal of Eye Movement Research\/}, {\bf 9}(1).

\bibitem[Brown {\em et~al.}(2020)Brown, Mann, Ryder, Subbiah, Kaplan, Dhariwal, Neelakantan, Shyam, Sastry, Askell, Agarwal, Herbert-Voss, Krueger, Henighan, Child, Ramesh, Ziegler, Wu, Winter, Hesse, Chen, Sigler, Litwin, Gray, Chess, Clark, Berner, McCandlish, Radford, Sutskever, and Amodei]{brown2020language}
Brown, T.~B., Mann, B., Ryder, N., Subbiah, M., Kaplan, J., Dhariwal, P., Neelakantan, A., Shyam, P., Sastry, G., Askell, A., Agarwal, S., Herbert-Voss, A., Krueger, G., Henighan, T., Child, R., Ramesh, A., Ziegler, D.~M., Wu, J., Winter, C., Hesse, C., Chen, M., Sigler, E., Litwin, M., Gray, S., Chess, B., Clark, J., Berner, C., McCandlish, S., Radford, A., Sutskever, I., and Amodei, D. (2020).
\newblock Language models are few-shot learners.

\bibitem[Chang {\em et~al.}(2021)Chang, Kong, Jia, and Meng]{chang2021chinese}
Chang, Y., Kong, L., Jia, K., and Meng, Q. (2021).
\newblock Chinese named entity recognition method based on bert.
\newblock In {\em 2021 IEEE International Conference on Data Science and Computer Application (ICDSCA)\/}, pages 294--299. IEEE.

\bibitem[Chiu and Nichols(2016)Chiu and Nichols]{chiu2016named}
Chiu, J.~P. and Nichols, E. (2016).
\newblock Named entity recognition with bidirectional lstm-cnns.
\newblock {\em Transactions of the association for computational linguistics\/}, {\bf 4}, 357--370.

\bibitem[Devlin {\em et~al.}(2019)Devlin, Chang, Lee, and Toutanova]{devlin2019bert}
Devlin, J., Chang, M.-W., Lee, K., and Toutanova, K. (2019).
\newblock Bert: Pre-training of deep bidirectional transformers for language understanding.

\bibitem[Escobar {\em et~al.}(2022)Escobar, Alcocer, Tarner, Beck, and Bergel]{escobar2022spike}
Escobar, R., Alcocer, J. P.~S., Tarner, H., Beck, F., and Bergel, A. (2022).
\newblock Spike--a code editor plugin highlighting fine-grained changes.
\newblock In {\em 2022 Working Conference on Software Visualization (VISSOFT)\/}, pages 167--171. IEEE.

\bibitem[Esuli and Sebastiani(2010)Esuli and Sebastiani]{esuli2010evaluating}
Esuli, A. and Sebastiani, F. (2010).
\newblock Evaluating information extraction.
\newblock In {\em Multilingual and Multimodal Information Access Evaluation: International Conference of the Cross-Language Evaluation Forum, CLEF 2010, Padua, Italy, September 20-23, 2010. Proceedings 1\/}, pages 100--111. Springer.

\bibitem[Face(2023)Face]{HuggingFace}
Face, H. (2023).
\newblock {BERT}.
\newblock \url{https://huggingface.co/docs/transformers/model_doc/bert}.
\newblock Accessed: 2023-08-23.

\bibitem[Gottipati {\em et~al.}(2011)Gottipati, Lo, and Jiang]{gottipati2011finding}
Gottipati, S., Lo, D., and Jiang, J. (2011).
\newblock Finding relevant answers in software forums.
\newblock In {\em 2011 26th IEEE/ACM International Conference on Automated Software Engineering (ASE 2011)\/}, pages 323--332. IEEE.

\bibitem[Hakala and Pyysalo(2019)Hakala and Pyysalo]{hakala2019biomedical}
Hakala, K. and Pyysalo, S. (2019).
\newblock Biomedical named entity recognition with multilingual bert.
\newblock In {\em Proceedings of the 5th workshop on BioNLP open shared tasks\/}, pages 56--61.

\bibitem[Hannebauer {\em et~al.}(2018)Hannebauer, Hesenius, and Gruhn]{hannebauer2018does}
Hannebauer, C., Hesenius, M., and Gruhn, V. (2018).
\newblock Does syntax highlighting help programming novices?
\newblock {\em Empirical Software Engineering\/}, {\bf 23}, 2795--2828.

\bibitem[He {\em et~al.}(2022)He, Xu, Yang, Han, Yang, and Lo]{He2022PTM4TagST}
He, J., Xu, B., Yang, Z., Han, D., Yang, C., and Lo, D. (2022).
\newblock Ptm4tag: Sharpening tag recommendation of stack overflow posts with pre-trained models.
\newblock {\em 2022 IEEE/ACM 30th International Conference on Program Comprehension (ICPC)\/}, pages 1--11.

\bibitem[Jehangir {\em et~al.}(2023)Jehangir, Radhakrishnan, and Agarwal]{jehangir2023survey}
Jehangir, B., Radhakrishnan, S., and Agarwal, R. (2023).
\newblock A survey on named entity recognition—datasets, tools, and methodologies.
\newblock {\em Natural Language Processing Journal\/}, {\bf 3}, 100017.

\bibitem[Jiang {\em et~al.}(2016)Jiang, Banchs, and Li]{jiang2016evaluating}
Jiang, R., Banchs, R.~E., and Li, H. (2016).
\newblock Evaluating and combining name entity recognition systems.
\newblock In {\em Proceedings of the Sixth Named Entity Workshop\/}, pages 21--27.

\bibitem[Kandpal {\em et~al.}(2022)Kandpal, Deng, Roberts, Wallace, and Raffel]{kandpal2022large}
Kandpal, N., Deng, H., Roberts, A., Wallace, E., and Raffel, C. (2022).
\newblock Large language models struggle to learn long-tail knowledge.
\newblock {\em arXiv preprint arXiv:2211.08411\/}.

\bibitem[Le~Guillarme and Thuiller(2022)Le~Guillarme and Thuiller]{le2022taxonerd}
Le~Guillarme, N. and Thuiller, W. (2022).
\newblock Taxonerd: deep neural models for the recognition of taxonomic entities in the ecological and evolutionary literature.
\newblock {\em Methods in Ecology and Evolution\/}, {\bf 13}(3), 625--641.

\bibitem[Li {\em et~al.}(2018a)Li, Li, Sun, Xing, Peng, Liu, and Zhao]{li2018improving}
Li, H., Li, S., Sun, J., Xing, Z., Peng, X., Liu, M., and Zhao, X. (2018a).
\newblock Improving api caveats accessibility by mining api caveats knowledge graph.
\newblock In {\em 2018 IEEE International Conference on Software Maintenance and Evolution (ICSME)\/}, pages 183--193. IEEE.

\bibitem[Li {\em et~al.}(2016)Li, Xing, Ye, and Zhao]{Li2016discussion}
Li, J., Xing, Z., Ye, D., and Zhao, X. (2016).
\newblock From discussion to wisdom: Web resource recommendation for hyperlinks in stack overflow.
\newblock In {\em Proceedings of the 31st Annual ACM Symposium on Applied Computing\/}, SAC '16, page 1127–1133, New York, NY, USA. Association for Computing Machinery.

\bibitem[Li {\em et~al.}(2018b)Li, Xing, and Sun]{Li2018LinkLiveDW}
Li, J., Xing, Z., and Sun, A. (2018b).
\newblock Linklive: discovering web learning resources for developers from q\&a discussions.
\newblock {\em World Wide Web\/}, {\bf 22}, 1699--1725.

\bibitem[Maity {\em et~al.}(2019)Maity, Panigrahi, Ghosh, Banerjee, Goyal, and Mukherjee]{maity2019}
Maity, S.~K., Panigrahi, A., Ghosh, S., Banerjee, A., Goyal, P., and Mukherjee, A. (2019).
\newblock Deeptagrec: A content-cum-user based tag recommendation framework for stack overflow.
\newblock In L.~Azzopardi, B.~Stein, N.~Fuhr, P.~Mayr, C.~Hauff, and D.~Hiemstra, editors, {\em Advances in Information Retrieval\/}, pages 125--131, Cham. Springer International Publishing.

\bibitem[Mireshghallah {\em et~al.}(2022)Mireshghallah, Uniyal, Wang, Evans, and Berg-Kirkpatrick]{mireshghallah2022memorization}
Mireshghallah, F., Uniyal, A., Wang, T., Evans, D., and Berg-Kirkpatrick, T. (2022).
\newblock Memorization in nlp fine-tuning methods.
\newblock {\em arXiv preprint arXiv:2205.12506\/}.

\bibitem[Nadeau and Sekine(2007)Nadeau and Sekine]{nadeau2007survey}
Nadeau, D. and Sekine, S. (2007).
\newblock A survey of named entity recognition and classification.
\newblock {\em Lingvisticae Investigationes\/}, {\bf 30}(1), 3--26.

\bibitem[Nadi and Treude(2020)Nadi and Treude]{nadi2020essential}
Nadi, S. and Treude, C. (2020).
\newblock Essential sentences for navigating stack overflow answers.
\newblock In {\em 2020 IEEE 27th International Conference on Software Analysis, Evolution and Reengineering (SANER)\/}, pages 229--239. IEEE.

\bibitem[Nguyen {\em et~al.}(2015)Nguyen, Wallace, and Lease]{nguyen2015combining}
Nguyen, A.~T., Wallace, B.~C., and Lease, M. (2015).
\newblock Combining crowd and expert labels using decision theoretic active learning.
\newblock In {\em Third AAAI conference on human computation and crowdsourcing\/}.

\bibitem[Overflow(????)Overflow]{SOEdit}
Overflow, S. (????).
\newblock {Edit a Question or Answer}.
\newblock \url{https://stackoverflowteams.help/en/articles/8858605-edit-a-question-or-answer }.
\newblock Accessed: 2024-04-01.

\bibitem[Overflow(2022)Overflow]{MarkdownHelp}
Overflow, S. (2022).
\newblock {Markdown help}.
\newblock \url{https://stackoverflow.com/editing-help}.
\newblock Accessed: 2023-01-30.

\bibitem[Palma {\em et~al.}(2022)Palma, Salza, and Gall]{palma2022fly}
Palma, M.~E., Salza, P., and Gall, H.~C. (2022).
\newblock On-the-fly syntax highlighting using neural networks.
\newblock In {\em Proceedings of the 30th ACM Joint European Software Engineering Conference and Symposium on the Foundations of Software Engineering\/}, pages 269--280.

\bibitem[Qiao {\em et~al.}(2020)Qiao, Li, Umer, and Guo]{qiao2020deep}
Qiao, L., Li, X., Umer, Q., and Guo, P. (2020).
\newblock Deep learning based software defect prediction.
\newblock {\em Neurocomputing\/}, {\bf 385}, 100--110.

\bibitem[Raffel {\em et~al.}(2020)Raffel, Shazeer, Roberts, Lee, Narang, Matena, Zhou, Li, and Liu]{t5}
Raffel, C., Shazeer, N., Roberts, A., Lee, K., Narang, S., Matena, M., Zhou, Y., Li, W., and Liu, P.~J. (2020).
\newblock Exploring the limits of transfer learning with a unified text-to-text transformer.
\newblock {\em The Journal of Machine Learning Research\/}, {\bf 21}(1), 5485--5551.

\bibitem[Ragkhitwetsagul {\em et~al.}(2019)Ragkhitwetsagul, Krinke, Paixao, Bianco, and Oliveto]{ragkhitwetsagul2019toxic}
Ragkhitwetsagul, C., Krinke, J., Paixao, M., Bianco, G., and Oliveto, R. (2019).
\newblock Toxic code snippets on stack overflow.
\newblock {\em IEEE Transactions on Software Engineering\/}, {\bf 47}(3), 560--581.

\bibitem[Ram{\'\i}rez {\em et~al.}(2019)Ram{\'\i}rez, Baez, Casati, and Benatallah]{ramirez2019understanding}
Ram{\'\i}rez, J., Baez, M., Casati, F., and Benatallah, B. (2019).
\newblock Understanding the impact of text highlighting in crowdsourcing tasks.
\newblock In {\em Proceedings of the AAAI Conference on Human Computation and Crowdsourcing\/}, volume~7, pages 144--152.

\bibitem[Rekha {\em et~al.}(2014)Rekha, Divya, and Bagavathi]{rekha2014}
Rekha, V.~S., Divya, N., and Bagavathi, P.~S. (2014).
\newblock A hybrid auto-tagging system for stackoverflow forum questions.
\newblock In {\em Proceedings of the 2014 International Conference on Interdisciplinary Advances in Applied Computing\/}, New York, NY, USA. Association for Computing Machinery.

\bibitem[Ren {\em et~al.}(2019)Ren, Xing, Xia, Li, and Sun]{ren2019discovering}
Ren, X., Xing, Z., Xia, X., Li, G., and Sun, J. (2019).
\newblock Discovering, explaining and summarizing controversial discussions in community q\&a sites.
\newblock In {\em 2019 34th IEEE/ACM International Conference on Automated Software Engineering (ASE)\/}, pages 151--162. IEEE.

\bibitem[Sarkar(2015)Sarkar]{sarkar2015impact}
Sarkar, A. (2015).
\newblock The impact of syntax colouring on program comprehension.
\newblock In {\em PPIG\/}, page~8.

\bibitem[Seaman(1999)Seaman]{seaman1999qualitative}
Seaman, C.~B. (1999).
\newblock Qualitative methods in empirical studies of software engineering.
\newblock {\em IEEE Transactions on software engineering\/}, {\bf 25}(4), 557--572.

\bibitem[Souza {\em et~al.}(2019)Souza, Nogueira, and Lotufo]{souza2019portuguese}
Souza, F., Nogueira, R., and Lotufo, R. (2019).
\newblock Portuguese named entity recognition using bert-crf.
\newblock {\em arXiv preprint arXiv:1909.10649\/}.

\bibitem[SPACY(2023)SPACY]{spacyNER}
SPACY (2023).
\newblock {Linguistic Features}.
\newblock \url{https://spacy.io/usage/linguistic-features#named-entities}.
\newblock Accessed: 2023-01-30.

\bibitem[StackExchange(2023)StackExchange]{StackExchangeMarkdown}
StackExchange (2023).
\newblock {How do I format my posts using Markdown or HTML?}
\newblock \url{https://meta.stackexchange.com/help/formatting}.
\newblock Accessed: 2023-01-30.

\bibitem[Strobelt {\em et~al.}(2015)Strobelt, Oelke, Kwon, Schreck, and Pfister]{strobelt2015guidelines}
Strobelt, H., Oelke, D., Kwon, B.~C., Schreck, T., and Pfister, H. (2015).
\newblock Guidelines for effective usage of text highlighting techniques.
\newblock {\em IEEE transactions on visualization and computer graphics\/}, {\bf 22}(1), 489--498.

\bibitem[Tanabe {\em et~al.}(2005)Tanabe, Xie, Thom, Matten, and Wilbur]{tanabe2005genetag}
Tanabe, L., Xie, N., Thom, L.~H., Matten, W., and Wilbur, W.~J. (2005).
\newblock Genetag: a tagged corpus for gene/protein named entity recognition.
\newblock {\em BMC bioinformatics\/}, {\bf 6}, 1--7.

\bibitem[Treude and Robillard(2016)Treude and Robillard]{treude2016augmenting}
Treude, C. and Robillard, M.~P. (2016).
\newblock Augmenting api documentation with insights from stack overflow.
\newblock In {\em 2016 IEEE/ACM 38th International Conference on Software Engineering (ICSE)\/}, pages 392--403. IEEE.

\bibitem[Viera {\em et~al.}(2005)Viera, Garrett, {\em et~al.}]{viera2005understanding}
Viera, A.~J., Garrett, J.~M., {\em et~al.} (2005).
\newblock Understanding interobserver agreement: the kappa statistic.
\newblock {\em Fam med\/}, {\bf 37}(5), 360--363.

\bibitem[Wang {\em et~al.}(2019a)Wang, Wang, Li, Xu, He, and Yang]{wang2019SOTagRec}
Wang, H., Wang, B., Li, C., Xu, L., He, J., and Yang, M. (2019a).
\newblock Sotagrec: A combined tag recommendation approach for stack overflow.
\newblock In {\em Proceedings of the 2019 4th International Conference on Mathematics and Artificial Intelligence\/}, page 146–152. Association for Computing Machinery.

\bibitem[Wang {\em et~al.}(2019b)Wang, Zhang, and Jiang]{Wang2019IEA}
Wang, L., Zhang, L., and Jiang, J. (2019b).
\newblock Iea: an answerer recommendation approach on stack overflow.
\newblock {\em Science China Information Sciences\/}, {\bf 62}(11), 212103.

\bibitem[Wang {\em et~al.}(2018)Wang, Lo, Vasilescu, and Serebrenik]{wang2018entagrec++}
Wang, S., Lo, D., Vasilescu, B., and Serebrenik, A. (2018).
\newblock Entagrec++: An enhanced tag recommendation system for software information sites.
\newblock {\em Empirical Software Engineering\/}, {\bf 23}, 800--832.

\bibitem[Wilson {\em et~al.}(2016)Wilson, Schaub, Ramanath, Sadeh, Liu, Smith, and Liu]{wilson2016crowdsourcing}
Wilson, S., Schaub, F., Ramanath, R., Sadeh, N., Liu, F., Smith, N.~A., and Liu, F. (2016).
\newblock Crowdsourcing annotations for websites' privacy policies: Can it really work?
\newblock In {\em Proceedings of the 25th International Conference on World Wide Web\/}, pages 133--143.

\bibitem[Wu and Yuan(2003)Wu and Yuan]{wu2003improving}
Wu, J.-H. and Yuan, Y. (2003).
\newblock Improving searching and reading performance: the effect of highlighting and text color coding.
\newblock {\em Information \& Management\/}, {\bf 40}(7), 617--637.

\bibitem[Xu {\em et~al.}(2017)Xu, Xing, Xia, and Lo]{xu2017answerbot}
Xu, B., Xing, Z., Xia, X., and Lo, D. (2017).
\newblock Answerbot: Automated generation of answer summary to developers' technical questions.
\newblock In {\em 2017 32nd IEEE/ACM International Conference on Automated Software Engineering (ASE)\/}, pages 706--716. IEEE.

\bibitem[Yang {\em et~al.}(2023)Yang, Wang, Li, and Wang]{yang2023does}
Yang, X., Wang, S., Li, Y., and Wang, S. (2023).
\newblock Does data sampling improve deep learning-based vulnerability detection? yeas! and nays!
\newblock In {\em 2023 IEEE/ACM 45th International Conference on Software Engineering (ICSE)\/}, pages 2287--2298. IEEE.

\bibitem[Ye {\em et~al.}(2016)Ye, Xing, Li, and Kapre]{ye2016software}
Ye, D., Xing, Z., Li, J., and Kapre, N. (2016).
\newblock Software-specific part-of-speech tagging: An experimental study on stack overflow.
\newblock In {\em Proceedings of the 31st Annual ACM Symposium on Applied Computing\/}, pages 1378--1385.

\bibitem[Yin {\em et~al.}(2019)Yin, Sun, Sun, and Jiao]{yin2019}
Yin, H., Sun, Z., Sun, Y., and Jiao, W. (2019).
\newblock A question-driven source code recommendation service based on stack overflow.
\newblock In {\em 2019 IEEE World Congress on Services (SERVICES)\/}, volume 2642-939X, pages 358--359.

\bibitem[Zhang {\em et~al.}(2019)Zhang, Wang, Chen, Zou, and Hassan]{zhang2019empirical}
Zhang, H., Wang, S., Chen, T.-H., Zou, Y., and Hassan, A.~E. (2019).
\newblock An empirical study of obsolete answers on stack overflow.
\newblock {\em IEEE Transactions on Software Engineering\/}, {\bf 47}(4), 850--862.

\bibitem[Zhang {\em et~al.}(2021a)Zhang, Wang, Chen, and Hassan]{zhang2021comments}
Zhang, H., Wang, S., Chen, T.-H., and Hassan, A.~E. (2021a).
\newblock Are comments on stack overflow well organized for easy retrieval by developers?
\newblock {\em ACM Transactions on Software Engineering and Methodology (TOSEM)\/}, {\bf 30}(2), 1--31.

\bibitem[Zhang {\em et~al.}(2021b)Zhang, Kang, Hooi, Yan, and Feng]{zhang2021deep}
Zhang, Y., Kang, B., Hooi, B., Yan, S., and Feng, J. (2021b).
\newblock Deep long-tailed learning: A survey.
\newblock {\em arXiv preprint arXiv:2110.04596\/}.

\bibitem[Zhu {\em et~al.}(2018)Zhu, Li, Conesa, and Pereira]{zhu2018gram}
Zhu, Q., Li, X., Conesa, A., and Pereira, C. (2018).
\newblock Gram-cnn: a deep learning approach with local context for named entity recognition in biomedical text.
\newblock {\em Bioinformatics\/}, {\bf 34}(9), 1547--1554.

\end{thebibliography}

\end{document}